# Monocular Vision-based Vehicle Localization Aided by Fine-grained Classification

Shuaijun Li, Yu Meng, Wei Li, Huihuan Qian and Yangsheng Xu

*Abstract*—Monocular camera systems are prevailing in intelligent transportation systems, but by far they have rarely been used for dimensional purposes such as to accurately estimate the localization information of a vehicle. In this paper, we show that this capability can be realized. By integrating a series of advanced computer vision techniques including foreground extraction, edge and line detection, etc., and by utilizing deep learning networks for fine-grained vehicle model classification, we developed an algorithm which can estimate vehicle's location (position, orientation and boundaries) within the environment down to 3.79% position accuracy and 2.5° orientation accuracy. With this enhancement, current massive surveillance camera systems can potentially play the role of e-traffic police and trigger many new intelligent transportation applications, for example, to guide vehicles for parking or even for autonomous driving.

## I. Introduction

Autonomous driving robots and vehicles have been researched for decades and still remain a hot topic [1]. One core aspect for autonomous driving is to equip robots or vehicles with the capability of self-localization, so that they can figure out where they are within the environment.

To achieve this, various sensors including GPS, cameras, Lidars and radars are installed [2]. For outdoor applications, GPS sensors are most frequently used for robot localization [3]. For indoor applications where GPS signals are too weak or even not available, other landmark based or map based approaches can also be used for localization [4]. For example: cameras can be used to look for artificial landmarks in the environment such as QR codes whose position were known in advance [5]; Lidar sensors can be used to obtain the local geometry features, and robot positions can be calculated by matching these local features in the global map [6].

All these approaches install sensors onto robots and figure out their localization by sensing the environment. Sensors can also be installed into the environment to do localization. Such configurations are commonly used in intelligent transportation system applications for vehicles or mobile robots [7]. The benefits are obvious: First, expensive sensors such as high resolution cameras and laser scanners can be installed onto the infrastructure and serve many robots or vehicles. Consequently, the requirements of intelligence for vehicles can be reduced, and overall cost of the intelligent system can be reduced as well. Secondly, sensors mounted in the environment usually possess a better overlook view and a larger field of view. This advantage can reduce the complexity and improve the accuracy of localization algorithms. For example in CES 2013, Audi used this type of configurations with Lidar installed in the garage to localize a vehicle and guide it for remote autonomous parking.

With regard to sensors installed in the environment, surveillance cameras are the most prevailing and thus promising ones, due to their low cost and massive existence. However by far, these monocular cameras are still mainly used for detection and recognition purposes, but rarely used for dimension-concerned purposes such as vehicle localization and vehicle position tracking [8]. Most of the current monocular vision-based vehicle tracking techniques only operate in the image level, while vehicles' positions in the world coordinate system cannot be accurately obtained yet.

Recently, some research has been focusing on using cameras for dimension-concerned purposes. Mrinal Haloi used videos from a monocular camera to estimate vehicle's relative position from road boundary in vehicle's width direction, and combined this with GPS data to obtain better localization accuracy [11]. In T.N. Schoepflin's research, they inversely transformed image coordinates into world coordinates and estimated vehicle speed based on vehicle's position estimations along the lane direction [12].

Strictly speaking, these works are not yet real vehicle localization solutions—which ideally would mean that the position, orientation and boundaries of the target vehicle can be accurately estimated. By far, there is no good solutions for this task yet, since it is still an open research regarding how to accurately estimate vehicles' dimensions (one pre-condition for localization).

For vehicle dimension estimation, modern computer vision solutions incorporate machine learning processes and transform this task into a classification problem. Various features are extracted and learned from training samples of vehicle datasets, and vehicle dimensions can be indirectly estimated by predicting the category (or model type) of the given vehicle image [13], [14]. The classification accuracy kept rising and has now reached over 90% with stimulus by

*This research is supported by the NSFC project U1613226 from the State Joint Engineering Lab and Shenzhen Engineering Lab on Robotics and Intelligent Manufacturing, China.

Shuaijun Li is with The Chinese University of Hong Kong, Hong Kong, China (e-mail: sjli01@mae.cuhk.edu.hk).
Huihuan Qian is with The Chinese University of Hong Kong, Shenzhen, Shenzhen, China (corresponding author, phone: +86 755-8427-3807; email: hhqian@cuhk.edu.cn).
Yu Meng, Wei Li and Yangsheng Xu are with The Chinese University of Hong Kong Shenzhen, Shenzhen, China (email:, mengyu139@126.com, 514072351@qq.com, ysxu@cuhk.edu.cn).

the annual ImageNet Large Scale Visual Recognition Challenge (ILSVRC) [17].

However, this coarse level classification only distinguish cars from other objects such as bicycles, (some research can distinguish a compact model from a mini model). From a dimension estimation point of view, a specific vehicle model such as Audi A4L needs to be recognized. Such problems are academically defined as fine-grained vehicle model classification [18], [20], whose accuracy is currently around 70%. With the development of deep neural networks [16], and with new structural datasets created for fine-grained car model classification such as CompCars [20], it becomes promising to further improve the accuracy to predict specific vehicle models and estimate the dimensions accordingly.

In this paper, we propose a monocular vision algorithm for vehicle localization (vehicle's position, orientation and boundaries). This algorithm properly applies a series of advanced CV techniques to obtain a vehicle's ROI image, coordinate transformation matrix and key footprint points. At the same time, this algorithm utilizes modern deep neural networks and new datasets such as CompCars to conduct fine-grained vehicle model classification and is able to predict a vehicle's model and estimate its dimensions accordingly. Eventually these results are combined to form the localization information of the vehicle.

The rest of the paper is organized as follow: Section II explains the proposed algorithm and its five major components with illustrations. In Section III, a series of experiments are conducted to validate the performance and robustness of the proposed algorithm, with detailed analysis to the test results. Conclusions are drawn and some future directions are discussed in Section IV.

## II. VEHICLE LOCALIZATION ALGORITHM DESIGN

### A. General Algorithm Flow Chart

Fig. 1 shows the general flow chart of the proposed algorithm. Video images are firstly obtained from monocular surveillance cameras, and then fed to two main algorithm pipelines in order to obtain the two key estimations: 1) vehicle dimensions and 2) key footprint points.

In the pipeline on the left, the vehicle ROI is first detected and extracted from video images using a GMM method [15] and shadow removal techniques. Subsequently, this vehicle ROI image is send to a deep learning network with the structure of GoogLeNet [16] for classification. This GoogLeNet is pre-trained on the ImageNet dataset [17] and fine-tuned on 431 car models in the CompCars dataset [20], so that it can be used for fine-grained vehicle model classification and is able to predict the specific vehicle model of the given ROI image. For example, to distinguish a BMW 7 Series vehicle from an Audi A4L is a fine-grained classification problem, rather than a coarse one as to tell a car from a bicycle. With the resulting vehicle model, the vehicle dimensions can be obtained from the database of vehicle specifications.

In the other pipeline on the right, the monocular camera is firstly calibrated to obtain the extrinsic parameters and the transformation matrix between image and world coordinates.

This is realized by detecting the corner of road lane markers using a series of computer vision techniques and measuring the coordinates of their corresponding points under the world frame. Meanwhile, the vehicle ROI image is subject to another series of visual processing including blob and cluster analysis, edge and line detection, etc. to detect the main footprint boundary lines. The license plate and wheels are also detected using available advanced computer vision algorithms [23], [24], and their image positions are incorporated in determining the key footprint points in the image. By further inverse transformation operations the world coordinates of the key footprint points can be estimated.

Finally, based on simple geometric relationships, the vehicle's orientation angle, center position and boundary positions can all be determined under world coordinates frame. This is how the vehicle's localization information are eventually obtained. Details of each process are fully explained in below.

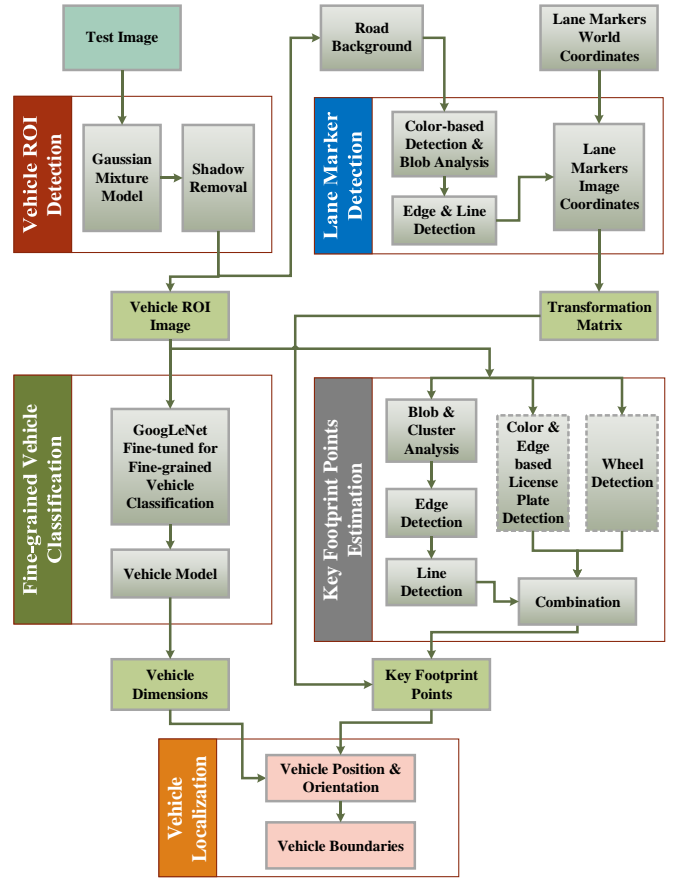

Figure 1 General Algorithm Flowchart

### B. Vehicle ROI Detection

First, we use Gaussian mixture model (GMM) method [15] to extract vehicle's region-of-interest (ROI) image as foreground from video streams. In GMM method, every pixel $X_t$ is modeled as a mixture of three Gaussians probability distributions over time, as shown in Eqn. (1). First 150 frames can be used to model the initial background, and new pixels that do not belong to any of the Gaussians distributions will be classified as foreground.

$$P(X_t) = \sum_{i=1}^{K} \omega_{i,t} N(X_t, \mu_{i,t}, c_{i,t}) \quad (1)$$

In the following updates after every foreground segmentation, by maximizing the expectation, the Gaussians' parameters such as their means, variances and weights can be estimated and will be updated adaptively to improve the detection and to allow the gradual changes in the scene. The final vehicle ROI detection results are shown in Fig. 2. Notice that we enlarge ROI region by 10% to each direction and use this enlarged image for the classification process in Sec. II-D to achieve better performance. We also applied a simple shadow removal algorithm [15] to polish the image for the following processing in Sec. II-E.

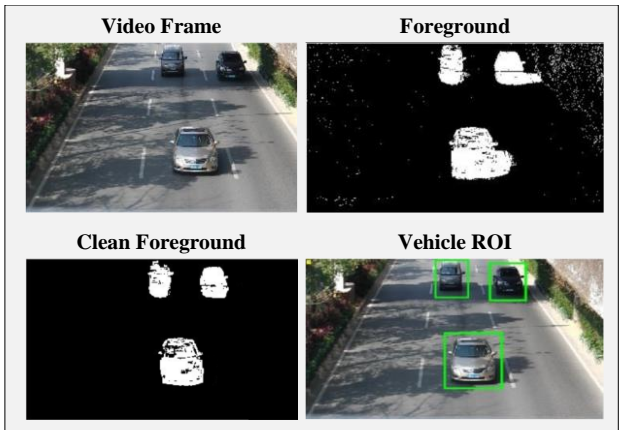

Figure 2 Vehicle ROI detection

### C. Lane Marker Detection and Camera Calibration

As the ultimate goal is to estimate the vehicle's localization in the global coordinates, the projective transformation matrix $H$ from the 2D world coordinates $X_W$ (in the road ground plane) to the 2D image coordinates $X_{Im}$ needs to be estimated (2). In other words, the monocular surveillance camera needs to be calibrated and its external parameters be determined.

$$X_{Im} = \begin{pmatrix} x_{Im} \\ y_{Im} \\ 1 \end{pmatrix} = H X_W = \begin{bmatrix} h_{11} & h_{12} & h_{13} \\ h_{21} & h_{22} & h_{23} \\ h_{31} & h_{32} & h_{33} \end{bmatrix} \begin{pmatrix} x_W \\ y_W \\ 1 \end{pmatrix} \quad (2)$$

Given enough image pixel points with known world coordinates $X_W$, this transformation matrix $H$ can be estimated. In this application of traffic surveillance, lane markers on the road can be used. These road lane markers are massive in existence and are of fixed color—either white or yellow. We can use basic computer vision techniques such as color-based ROI extraction and blob analysis to detect these ROIs. Subsequently, by applying canny edge detection and hough transform line detection, the corner points of the road lane markers can be successfully detected and their image coordinates can be found.

When the world frame has been properly established, world coordinates of these lane markers can be measured manually or from accurate maps. By using RANSAC algorithm, outliers can be eliminated and the transformation matrix $H$ can be estimated. By applying the inverse transformation $H^{-1}$, the world coordinates of any image pixels can be determined. Fig. 3 shows the detection process for lane markers and the map under world coordinates frame after the inverse transformation. This $H^{-1}$ will be used in the later process to obtain the world coordinates for footprint points in Sec. II-E.

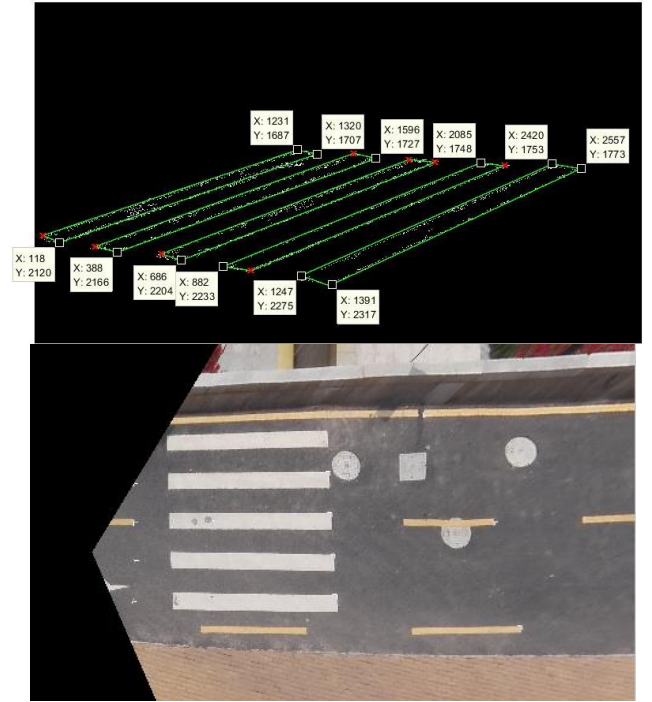

Figure 3 Camera Calibration based on Lane Marker Detection

### D. Fine-grained Vehicle Classification with Deep Learning

In this step, the vehicle ROI image provided by Sec. II-B is further subject to a fine-grained vehicle model classification process. Research regarding vehicle classification has been around for decades. This issue was firstly addressed as one type of object classification task. Stimulated by the annual ImageNet Large Scale Visual Recognition Challenges (ILSVRC), and strongly influenced by the adoption of deep learning neural networks, the overall top-five error rate of object classification has now reached around 7% by VGG and GoogLeNet networks in 2014 [17].

However, this result refers to vehicle classification on a coarse level, for example, to distinguish a passenger car image from a bicycle image. To distinguish a Audi A4 from a BMW 7 is more appropriately defined as a fine-grained vehicle model classification problem. Starting from 2013, some research interest has shifted to this fine-grained image classification task [18], [19]. A good dataset and a suitable learning model are the two critical elements to achieve a good classification accuracy. In this paper, we utilize the CompCars dataset [20] and the GoogLeNet network structure for the task of fine-grained vehicle model classification.

*1) CompCars Dataset*

The Comprehensive Cars (CompCars) dataset was created in 2015 by researchers from The Chinese University of Hong Kong (CUHK). This dataset contains over 1600 car models from over 160 car makes, covering most of the commercial passenger vehicle models in the recent decade. It contains 136,726 images of entire-car from various view angles and 27,618 images of car parts. Most importantly, the images are hierarchically labeled in Car Make, Car Model and Year of

Manufacturing as shown in Fig. 4. This property gives this dataset unique advantage for fine-grained vehicle model classification [20].

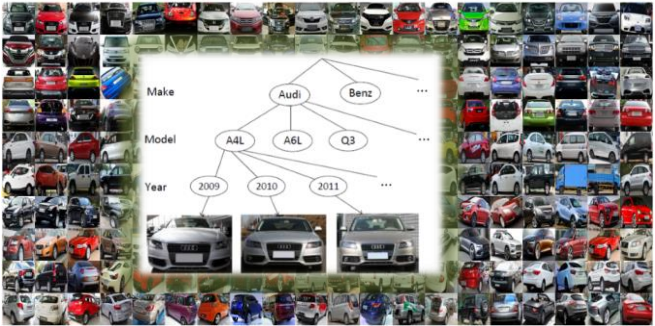

Figure 4 CompCars Dataset

*2) Network Structure – GoogLeNet*

Efforts has been made using CompCars for fine-grained vehicle model classification using Overfeat model [20], and the accuracy was 76.7% for Top-1 and 91.7% for Top-5. This accuracy has to be further improved for our task of vehicle dimension estimation, and a natural direction for improvement is the network structure.

With the regard to the model structures, some research are trying to develop extra network to detect and classify vehicle parts [19] to aid with fine-grained classification. And some other research followed the direction of making the network deeper. GoogLeNet, the winner of ILSVRC 2014, is such a deep network. It introduced the innovative Inception Module and cascade this module to form its network as shown in Fig. 5. It was recently reported that, when CompCars datasets are fed into GoogLeNet with parameters pre-trained from ImageNet for fine-tuning. This model can obtain a top-1 accuracy 91.2% and a top-5 accuracy 98.1% on the testing set. Given these promising background, we choose to use the GoogLeNet network structure and use the fine-tuned parameters in our algorithm for fine-grained vehicle model classification.

| type | patch size/ stride | output size | depth | #1×1 | #3×3 reduce | #3×3 | #5×5 reduce | #5×5 | pool proj | params | ops |
|---|---|---|---|---|---|---|---|---|---|---|---|
| convolution | 7×7/2 | 112×112×64 | 1 | | | | | | | 2.7K | 34M |
| max pool | 3×3/2 | 56×56×64 | 0 | | | | | | | | |
| convolution | 3×3/1 | 56×56×192 | 2 | | 64 | 192 | | | | 112K | 360M |
| max pool | 3×3/2 | 28×28×192 | 0 | | | | | | | | |
| inception (3a) | | 28×28×256 | 2 | 64 | 96 | 128 | 16 | 32 | 32 | 159K | 128M |
| inception (3b) | | 28×28×480 | 2 | 128 | 128 | 192 | 32 | 96 | 64 | 380K | 304M |
| max pool | 3×3/2 | 14×14×480 | 0 | | | | | | | | |
| inception (4a) | | 14×14×512 | 2 | 192 | 96 | 208 | 16 | 48 | 64 | 364K | 73M |
| inception (4b) | | 14×14×512 | 2 | 160 | 112 | 224 | 24 | 64 | 64 | 437K | 88M |
| inception (4c) | | 14×14×512 | 2 | 128 | 128 | 256 | 24 | 64 | 64 | 463K | 100M |
| inception (4d) | | 14×14×528 | 2 | 112 | 144 | 288 | 32 | 64 | 64 | 580K | 119M |
| inception (4e) | | 14×14×832 | 2 | 256 | 160 | 320 | 32 | 128 | 128 | 840K | 170M |
| max pool | 3×3/2 | 7×7×832 | 0 | | | | | | | | |
| inception (5a) | | 7×7×832 | 2 | 256 | 160 | 320 | 32 | 128 | 128 | 1072K | 54M |
| inception (5b) | | 7×7×1024 | 2 | 384 | 192 | 384 | 48 | 128 | 128 | 1388K | 71M |
| avg pool | 7×7/1 | 1×1×1024 | 0 | | | | | | | | |
| dropout (40%) | | 1×1×1024 | 0 | | | | | | | | |
| linear | | 1×1×1000 | 1 | | | | | | | 1000K | 1M |
| softmax | | 1×1×1000 | 0 | | | | | | | | |

Figure 5 GoogLeNet Network Structure

*3) Test Results on Real-world Vehicle Images*

Due to the complex real-world environment, the vehicle ROI image we obtained in Sec. II-A can be very different from test images in ImageNet or even CompCars. A series of tests and investigations were conducted to validate the classification accuracy and to finally estimate vehicle dimensions.

First, we randomly shot some vehicle images in real world around our research institute covering 10 types of vehicle models, and we extracted vehicle's ROI accordingly and sent them into GoogLeNet for classification. It turned out that all of their models are correctly predicted with 100% accuracy. We also closely observed the probability of the Top-5 predictions, as shown in Fig. 6. The Top-1 prediction is winning by a large ratio, indicating a good algorithm robustness.

| BMW 7 | Benz R Class | Citroen C4L | Elantra Yuedong |
|---|---|---|---|
| BMW 7 | Benz R Class | Citroen C4L | Elantra Yuedong |
| BMW 5 | Sonata | Sharan | Sunshine |
| BMW 5 GT | Lexus LS | Citroen C5 | Permacy |
| BMW 3 | Peugeot 508 | Roewe 350 | Jingyi |
| BMW 3 Convertible | Cadillac CTS | C-Elysee sedan | Sonata |

Figure 6 Fine-grained Vehicle Model Classification Top-5 Result

However, this result can be too optimistic. To further test the algorithm robustness, we tested the algorithm more exaggeratedly with more diversified real-world vehicle ROI images. As shown in Fig. 7, we collected 10 different vehicle models and 10 ROI images for each model from the internet. These images contain diversified road background which is typical for real-world traffic surveillance but are not typical for CompCars dataset images.

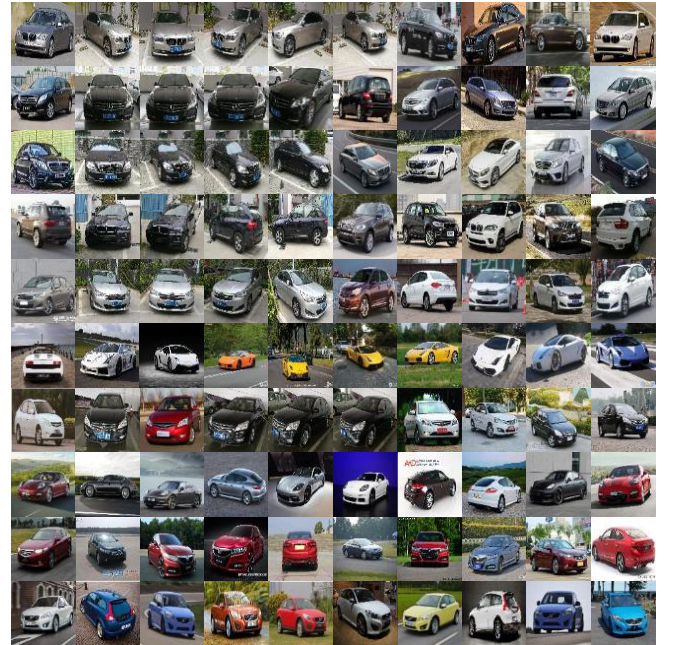

Figure 7 Test Vehicle ROI Images of 10 Models

Test results are listed in Table 1. In the middle portion, we recorded the prediction ranks of the actual vehicle model. For example, in row 2, Benz R Series vehicle images are mostly classified accurately as Top 1 prediction, but the 2nd test image is classified wrongly as other vehicle models, say Benz S Series, while the correct vehicle model of Benz R Series is ranking 2 out of 5 in the Top-5 predictions of the algorithm. It can be seen it is not guaranteed yet that the vehicle model can always be predicted correctly, but those wrong predictions are not too far away, so maybe we can still make good use of these predictions.

Table 1 Test Results and Dimension Estimation Errors

|    | Vehicle Model | Top #/5 as Classification Result | | | | | | | | | | Actual Vehicle Dimensions | | | Average Errors (%) | | |
|----|---------------|---|---|---|---|---|---|---|---|---|----|------|------|------|-------|-------|-------|
|    |               | 1 | 2 | 3 | 4 | 5 | 6 | 7 | 8 | 9 | 10 | L    | W    | H    | L     | W     | H     |
| 1  | BMW 7 Series  | 1 | 1 | 1 | 1 | 1 | 1 | 3 | 1 | 1 | 1  | 5250 | 1902 | 1498 | 0.57% | 0.16% | 1.84% |
| 2  | Benz R Series | 1 | 2 | 1 | 1 | 1 | 1 | 1 | 1 | 1 | 1  | 5159 | 1922 | 1663 | 0.66% | 0.45% | 1.13% |
| 3  | Benz S Series | 2 | 1 | 1 | 1 | 1 | 1 | 3 | 1 | 1 | 2  | 5250 | 1899 | 1494 | 0.94% | 0.69% | 0.36% |
| 4  | BMW X5        | 1 | 1 | 1 | 1 | 1 | 1 | 1 | 1 | 1 | 1  | 4909 | 1938 | 1772 | 0.00% | 0.00% | 0.00% |
| 5  | Citroen C4L   | 1 | 1 | 1 | 1 | 1 | 1 | 1 | 1 | 1 | 1  | 4675 | 1780 | 1500 | 0.00% | 0.00% | 0.00% |
| 6  | Gallardo      | 1 | 1 | 1 | 1 | 1 | 5 | 1 | 1 | 1 | 1  | 4345 | 1900 | 1165 | 0.42% | 0.25% | 1.02% |
| 7  | Hyundai_Elantra | 1 | 1 | 1 | 1 | 1 | 1 | 1 | 1 | 1 | 1 | 4545 | 1725 | 1425 | 0.00% | 0.00% | 0.00% |
| 8  | Panamera      | 1 | 1 | 1 | 1 | 1 | 1 | 1 | 1 | 1 | 1  | 5049 | 1937 | 1423 | 0.00% | 0.00% | 0.00% |
| 9  | Spirior       | 1 | 1 | 1 | 1 | 1 | 1 | 1 | 1 | 1 | 1  | 4840 | 1850 | 1465 | 0.00% | 0.00% | 0.00% |
| 10 | Volvo_C30     | 3 | 1 | 1 | 1 | 1 | 1 | 2 | 1 | 1 | 1  | 4266 | 1782 | 1447 | 1.20% | 0.61% | 0.47% |

Our idea is, given that this network can make wrong predictions at times, it is indeed only confused by some vehicle models who are very similar to the tested model. These predicted similar-but-confusing vehicle models are probably also of similar vehicle dimensions with the tested model as well. So we can still use these "wrong predictions" for vehicle dimension estimations.

Based on this idea, we choose to always trust the network's prediction results and use the vehicle dimensions of the predicted model as our vehicle dimension estimations. And we treat the differences between the predicted model and the actual model as estimation errors. It can be seen in the right portion of Table 1, for some cases in which vehicle models are always predicted correctly, the average dimension estimation errors are 0; for other cases in which vehicle models are occasionally predicted incorrectly, the average dimensions estimation errors are less than 1.5% for length and width dimensions and less than 2% for height dimension. These approximately correspond to 50 mm error for length, 10 mm error for width and 25 mm error for height. This accuracy of vehicle dimension estimations are further tested on real-world traffic videos, and the details will be introduced in Sec. III-A.

Based on these results, we believe these dimension estimation errors are acceptable, because compared with footprint estimation errors to be addressed in Sec. II-E, these dimension errors are of less weight, and thus having a less impact on the final localization estimation errors. So in our algorithm design, we choose to always trust these "wrong model predictions" in a "dimensionally-correct" sense.

*E. Key Footprint Points Estimation*

Now we have the vehicle dimension information from Sec. II-D, and it is represented under world frame like a boundary box with known dimensions from vehicles' specifications. To actually localize the vehicle, we need one more step to determine some key footprint points for the boundary box to stick to. To realize this, in this process as shown in Fig. 1, we mainly apply a series of computer vision techniques to detect the main boundary lines of the vehicle, and we also detect vehicle's license plate or wheels in parallel, and finally, these information are combined to determine these key footprint points.

Frist, we extract vehicle's main frame lines and determined the footprint lines based on the same framework in [22]. Concretely, with the given vehicle ROI image, blob analysis and clustering analysis is firstly applied to remove the high frequency component mainly in the car body's color regions [21]. After that a canny edge detection process is applied to extract vehicle's edge pixels and followed by a hough-transform line detection to detect the major lines of the vehicle body. Notice that similar angle filtering are applied as in [22] to exclude other unwanted lines in other directions. As the result, as shown in Fig. 8, several side lines in vehicle's length directions denoted as $l_S$ and several lines in vehicle's width direction denoted as $l_W$ can be detected. Subsequently, we average the info of all width lines $l_{W1-N}$ and extrapolate it to the bottom margin of the ROI by tangency to form the front line of footprint, denoted as $l_F$. We conduct similar operations on all the side lines $l_{S1-N}$ to form the side line of footprint, denoted as $l_S$. Based on these two footprint lines of $l_F$ and $l_S$, and aided by detected license plate or wheels positions, key footprint points can be determined.

Secondly, also based on the vehicle ROI image, we apply a series of color-and-edge based license plate detection algorithm in parallel [23]. Same process as in [22] is applied to detect the region of license plate based on transformed HSV info first, and then the upper and lower lines (denoted as $l_{UL}$ and $l_{LL}$) of vehicle's license plate can be detected using edge and line detection algorithms. The middle point of license plate $P$ is selected to aid with the following combination process.

Thirdly, we use one more branch of algorithm to detect possible vehicle wheels from the ROI image. To do this, the 2D ellipses matching method and algorithm processes as introduced in [24] are implemented. Details are omitted here due to the space limit. When wheels are indeed detected, their crossing points with $l_S$ will be determined as the tire positions and the middle point of this two tire positions $Q$ will be used to aid with the following combination process.

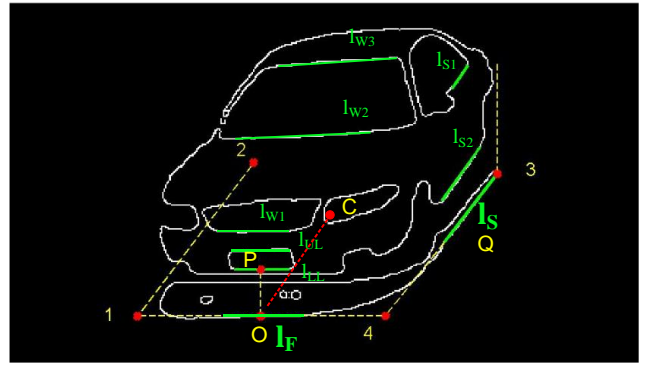

Figure 8 Key Footprint Points Estimation based on Footprint Line's Orientations and License Plate's Positions

Finally in the combination process, the key foot print points are determined as follows: When license plate is indeed detected, point $O$ will be used as the key point for position localization. While, the angular direction of $l_S$ will be used for orientation localization. This is achieved by drawing a parallel line to $l_S$ from point $O$ in the image, and pick a point $C$ at a fixed distance along this parallel line. The angular direction of $OC$ will be considered as vehicle's length orientation. It is worth to mention that, when license plate are detected, vehicle's front or rear side is considered to be detected, and this point $O$ and $C$ will be used as the key points with the highest priority. For some other cases, due to different pose angles, it may be difficult to see the vehicle's front side and license plate, we then detect vehicle wheels instead. In such circumstances, the point $Q$ will be used as the key point for position localization. And a similar $QC$ will be generated using $l_F$ as reference parallel line. Points $Q$ and $C$ will be used only if license plate cannot be detected from the ROI image.

### F. Vehicle Localization

With the key footprint points determined in the image, say point $O_{im}$ and $C_{im}$ aided by license plate detection. The world coordinates of these key points $O_W$ and $C_W$ can be calculated by applying the inverse transformation process of $H^{-1}$ in Eqn. (2). We have a boundary box with known dimensions from to be localized, so we stick the corresponding point (the middle point of vehicles' frontline calculated from specification dimensions) of this boundary box to this point of $O_W$. And we align the longer centerline of the boundary box with the line of $O_W C_W$. Regarding the case of having $Q_{im}$ and $C_{im}$ as key footprint points, the same process will be used, but instead, the middle point of vehicle's two wheel axles will be used as the corresponding point and the short centerline of the boundary box will be used for orientation alignment.

Fig. 9 shows the final localization result (drawn in green) under the world coordinates frame. Coordinates of the vehicle's center position, boundaries positions and orientation angles are all estimated now, realizing the ultimate goal of the proposed algorithm. When there is a way to obtain the real localization information, these estimated vehicle localization can be compared with the real localization for accuracy analysis, as shown in Fig. 9. We will address this issue in Sec. III-B.

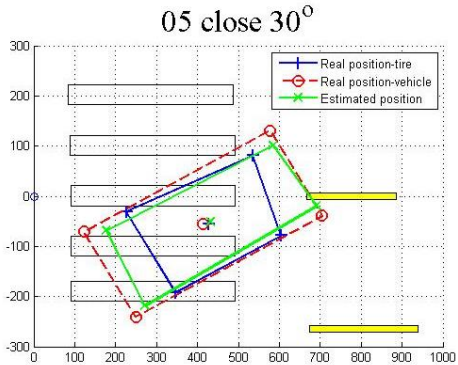

Figure 9 Final Localization Result Compared with Real Info

### III. VALIDATION TESTS AND EXPERIMENTS

#### A. Test on Real-world Traffic Videos

To test the performance of the proposed algorithm, we conducted more tests for validation. We firstly tried to test the algorithm performance on some real-world traffic videos.

We manually configured a surveillance camera above the road to have a proper field of view, and several videos with 720P resolutions were recorded for tests. As shown in Fig. 10, in real-world videos, background can be complex and may contain some shadows, etc., and vehicles can be driving at different speed.

Test results (similar to Table 1 not omitted here) showed that 60% of the vehicles' models can be correctly classified while the others were incorrectly predicted as other models and thus inducing some error for the following dimensions estimations. On average the errors are 1.9% or 87.8 mm for length, 2.5% or 45.8 mm for width, 3.2% or 47.7 mm for height. These errors are larger than the results we tested earlier in Sec. II-D, so we claim the dimension estimation accuracy to be 2.5% for length and width to be conservative.

These vehicle dimensions can be further transformed to coordinates under world frame as localization estimates, but it is difficult to obtain with their real localization information for comparison and accuracy analysis. So we arranged another test in Sec. III-B to deal with this concern.

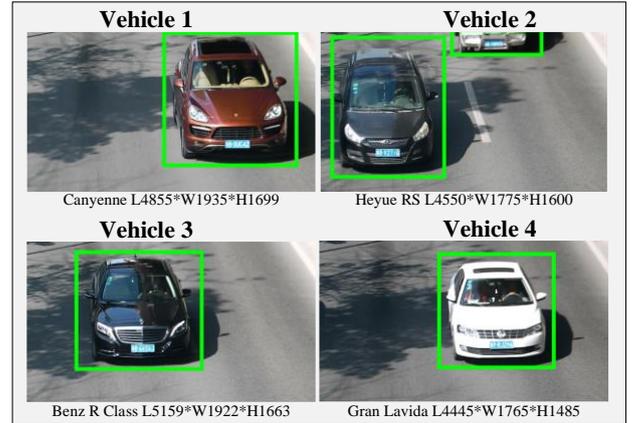

Figure 10 Real-world Traffic Test Video Frames

The proposed algorithm was written in Python under Linux environment with Caffe and openCV libraries and interfaces, we also tried two types of hardware configurations: 1) a Lenovo laptop with i5 CPU and 8G RAM, 2) a HP Z240 workstation with Nvidia GTX1070 GPU with 8G GDDR5, i7 CPU and 16G RAM. In the former configuration, the entire algorithm takes about 2 seconds, while in the later, it takes only 0.1 second and can be considered as real-time processing.

### B. Test for Localization Accuracy Analysis

In order to analyze the localization accuracy of the proposed algorithm, we conducted another test in which the real vehicle localization can be obtained for comparison.

#### 1) Experiment Setup

First, we chose a suitable piece of testing road with lane markers and setup a surveillance camera at a fixed position. We then established a world coordinates frame with its origin in the middle of the road, X axes across the road pointing to the right and Y axes along the road. With this, the world coordinates of these lane markers can be measured to aid transformation estimations in Sec. II-C, Fig. 3.

Within this tests environment, a test vehicle was driven to different positions with different orientations and was stopped for a while, as shown in Fig. 13. During each stop, the positions of vehicle's four tires were marked as shown in Fig. 11 and their world coordinates be measured afterwards. Note that all the geometric dimensions and relationships of the test vehicle including wheelbase can be obtained from vehicle specifications, so that based on these four tires' positions, coordinates for vehicle's center position, boundaries positions and orientation angle can all be calculated, as shown in Fig. 9.

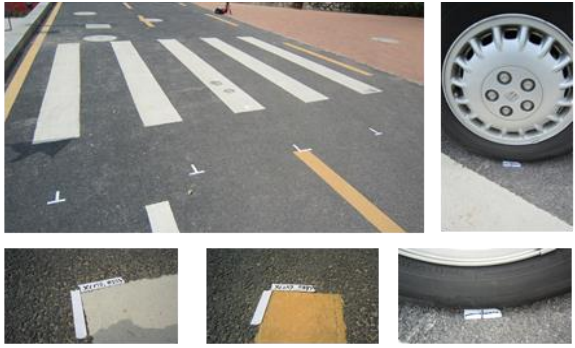

Figure 11 Test Setup and Position Markers

Now the real localization information has been obtained, and we compare it with the estimated localization from our proposed algorithm in the same figure for accuracy analysis. As shown in Fig. 9 and Fig. 13. The red lines with cycle markers represents the vehicle's real localization coordinates. The blue points represents the four tires' positions marked in the tests, and the red points are calculated based on these blue points. The green lines with cross markers represent the vehicle's estimated localization by our proposed algorithm. We tested six different configurations in which test vehicles are of three various pose angles of -30°, 0° and 30° to the X axes, and different distances to the camera, as shown in Fig. 13. The test results and are listed in Table 2 for accuracy analysis.

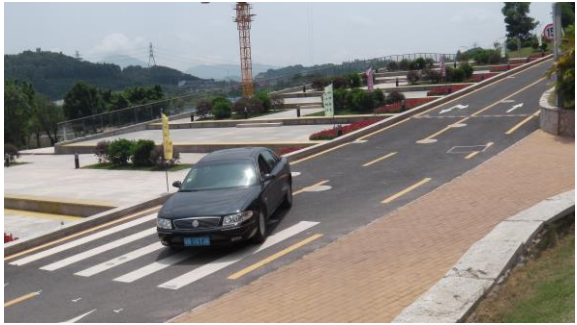

Figure 12 Test Video Frame for Localization Accuracy Analysis

*2) Experimental Results*

The testing results show that the proposed algorithm can successfully estimate vehicle localization with a large variety of posing angles and distances. Particularly, this algorithm remains valid at some extreme vehicle pose angles close to pure front-view or pure side-view. This is apparently an advantage over some other direct dimension estimation approaches based on geometric features.

As shown in Table 2, when localizing vehicles with different pose angles and distances, the orientation error is less than 2.5 degrees. The position estimation error is less than 120 mm or 2.5% in X direction and less than 50 mm or 2.5% in Y direction. This will jointly generate an estimation error of 150 mm, and compared with the size of the vehicle of 4920 by 1870 mm, this equals to a positioning accuracy of 2.85%. This positioning accuracy applied to all of vehicle's center and boundary positions. In this test, the vehicle model was correctly classified, if we include the dimension estimation errors of 2.5% in Sec. III-A, the algorithm's final localization accuracy will be 3.79%.

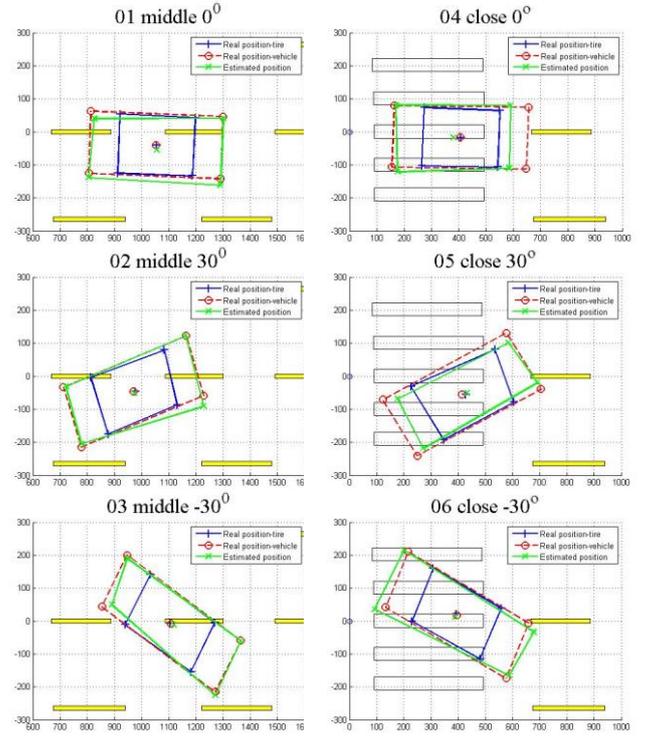

Figure 13 Tests with Different Vehicle Positions and Orientations

Table 2 Test Results Summary

| Real Vehicle Localization | Position X (mm) | Position Y (mm) | Orientation Angle (°) | |
|---|---|---|---|---|
| Middle 0 | 404 | 10530 | -3.76 | |
| Middle 30 | 456 | 9706 | 16.95 | |
| Middle -30 | 76 | 11087 | -32.71 | |
| Close 0 | 194 | 4060 | -2.07 | |
| Close 30 | 548 | 4147 | 23.92 | |
| Close -30 | -181 | 3958 | -26.30 | |
| **Estimated Vehicle Localization** | *Position X (mm)* | *Position Y (mm)* | *Orientation Angle (°)* | |
| Middle 0 | 286 | 10499 | -1.31 | |
| Middle 30 | 493 | 9659 | 16.55 | |
| Middle -30 | 155 | 11131 | -31.62 | |
| Close 0 | 121 | 4043 | -1.59 | |
| Close 30 | 478 | 4159 | 22.08 | |
| Close -30 | -143 | 3915 | -24.80 | |
| **Estimation Error** | *X Error (mm)* | *Y Error (mm)* | *Angle Error (°)* | *Position Error (mm)* |
| Middle 0 | -118 | -31 | 2.45 | 122.00 |
| Middle 30 | 37 | -47 | -0.40 | 59.56 |
| Middle -30 | 79 | 44 | 1.09 | 90.43 |
| Close 0 | -73 | -17 | 0.48 | 75.27 |
| Close 30 | -70 | 13 | -1.84 | 71.11 |
| Close -30 | 38 | -43 | 1.50 | 57.52 |
| **Estimation Error%** | *X Error (%)* | *Y Error(%)* | *Angle Error(%)* | *Position(%)* |
| Middle 0 | 2.40% | 1.66% | / | 2.32% |
| Middle 30 | 0.75% | 2.49% | / | 1.13% |
| Middle -30 | 1.61% | 2.35% | / | 1.72% |
| Close 0 | 1.49% | 0.91% | / | 1.43% |
| Close 30 | 1.42% | 0.67% | / | 1.35% |
| Close -30 | 0.78% | 2.30% | / | 1.09% |

## IV. Conclusion And Future work

To conclude, in this paper, an algorithm is proposed to estimate a vehicle's localization information including center position, orientation and boundary positions from monocular camera images.

The algorithm using GMM method to extract the vehicle's ROI and send it to a deep learning neural network with the GoogLeNet structure to predict the vehicle's specific model type. The dimensions of the predicted vehicle model is used as the estimations to the given vehicle. Tests showed that the dimension estimations accuracy is less than 2.5% for vehicle's length and width.

The algorithm also applies a series of computer vision techniques to detect the vehicle's main frame lines, license plate, wheels, and eventually combines all these detection to determine key footprint points in the image. With the calibrated camera parameters, these key points can be inversely transformed back to world coordinates frame. By adding the boundary box with estimated vehicle dimension on top of these key footprint points, all of vehicle's localization information are obtained. Tests showed that the final localization accuracy is 3.79% for position and 2.5° for orientation.

This research suggests that it is possible to conduct real-time vehicle localization using monocular surveillance camera systems down to reasonable accuracy. Based on this work, current massive surveillance camera can be equipped with this primary localization capability and many potential applications can be triggered off. With the advantage of the better and larger field-of-view, and the potential of multiple object localization in real-time, this technology may provide great help to solve some of the challenging intelligent transportation problems by aiding those intelligent vehicles with only ego sensors.

For example, such intelligent camera systems can be used to guide the dense traffic at road crossings, to assist vehicles for parking within complicated parking lot, or even to guide vehicles for autonomous driving within a camera-monitored community. The next step of our research is to develop such a demonstration system, in which intelligent vehicles with low-cost ego sensors can be guided by this kind of intelligent camera systems within a community for autonomous driving and parking.

## References

[1] T. Luettel, et al., "Autonomous Ground Vehicles - Concepts and a Path to the Future," *Proc. IEEE*, vol.100, 2012, pp. 1831-1839.
[2] C. Ilas, "Electronic sensing technologies for autonomous ground vehicles: A review," *Int. Symp. Adv. Topics in Elect. Eng. IEEE*, 2013, pp. 1–6.
[3] A. Almeida and O. Khatib, *Autonomous Robotic Systems*, London, UK, Springer London, 1998.
[4] R. Negenborn, "Robot Localization and Kalman Filters," M.S. thesis, Inst. Inform. Comput. Sci., Utrecht Univ., Netherlands, 2003.
[5] T. Suriyon et al., "Development of Guide Robot by Using QR Code Recognition," in *2nd TSME Int. Conf. Mech. Eng.*, Krabi, 2011
[6] X. Yuan et al., "Lidar Scan-Matching for Mobile Robot Localization," *Inform. Tech. J.,* vol. 9, no. 1, pp. 27–33, 2010.
[7] S. Sivaraman et al., "Looking at Vehicles on the Road_A Survey of Vision-Based Vehicle Detection, Tracking, and Behavior Analysis," *IEEE Trans. Intell. Transp. Syst. (ITS)*, vol. 4, pp.1773–1795, 2013.
[8] B. Tian et al., "Hierarchical and Networked Vehicle Surveillance in ITS: A Survey," *IEEE Trans. Intell. Transp. Syst. (ITS),* vol. 16, no. 2, pp. 557–580, 2015.
[9] R. Ratajczak et al., "Vehicle size estimation from stereoscopic video," in *19th Int. Conf. Sys. Signl. Img. Proc. (IWSSIP)*, Vienna, 2012, pp. 405–408.
[10] R. Ratajczak et al., "Vehicle Dimensions Estimation Scheme Using AAM on Stereoscopic Video," *IEEE Int. Conf. Adv. Video Sign. Survl. (AVSS)*, Krakow, Poland, 2013, pp. 478–482.
[11] M. Haloi and D. Jayagopi, "Vehicle Local Position Estimation System," *IEEE Int. Conf. Veh. Electron. Saf.*, Hyderabad, India, 2015.
[12] T.N. Schoepflin and D.J. Dailey, "Dynamic camera calibration of roadside traffic management cameras for vehicle speed estimation," *IEEE Trans. Intell. Transp. Syst. (ITS)*, vol. 4, no. 2, pp. 90–98, 2003.
[13] Buch, N. et al., "A Review of Computer Vision Techniques for the Analysis of Urban Traffic," *IEEE Trans. Intell. Transp. Syst. (ITS),* vol. 12, no. 3, pp. 920–939, 2011.
[14] Saran K B and Sreelekha G, "Traffic video surveillance: Vehicle detection and classification," *Int. Conf. Control Commun. Comput. (ICCC),* India, 2015, pp. 516–521.
[15] Zeiler et al., "Visualizing and Understanding Convolutional Networks," *European Conf. Comput. Vision (ECCV)*, vol. 8689, pp. 818-833, 2014.
[16] C. Szegedy et al., "Going deeper with convolutions," *IEEE Conf. Comput. Vision Pattern Recognition (CVPR)*, Boston, MA, pp.1–9, 2015,
[17] Russakovsky et al., "ImageNet Large Scale Visual Recognition Challenge," *Int. J. Comput. Vision, vol.* 115, no.3 , pp. 211–252, 2015.
[18] F. Zhou and Y. Lin, "Fine-Grained Image Classification by Exploring Bipartite-Graph Labels," *IEEE Conf. Comput. Vision Pattern Recognition (CVPR)*, Las Vegas, NV, pp. 1124-1133, 2016.
[19] Krause, Jonathan et al. "Learning Features and Parts for Fine-Grained Recognition," *IEEE Conf. Comput. Vision Pattern Recognition (CVPR)*, Stockholm, 2014, pp. 26–33.
[20] L. Yang et al., "A large-scale car dataset for fine-grained categorization and verification," *IEEE Conf. Comput. Vision Pattern Recognition (CVPR)*, Boston, MA, 2015, pp. 3973–3981.
[21] Wu et al., "Vehicle Orientation Detection Using Vehicle Color and Normalized Cut Clustering," *Int. J. Pattern Recognition and Artificial Intell. (IJPRAI),*vol.24, no. 5, pp. 823-846, 2009.
[22] Shuaijun Li et al, "Vehicle 3-Dimension Measurement By Monocular Camera Based on License Plate," presented at *IEEE Int. Conf. Robotics and Biomimetics (ROBIO)*, Qingdao, 2016.
[23] Gilly et al., "A Survey on License Plate Recognition Systems," *Int. J. Comp. Applications* vol. 61, no. 6, pp. 34–40, 2013.
[24] Hutter M and N. Brewer. "Matching 2-D ellipses to 3-D circles with application to vehicle pose identification," *Int. Conf. Image and Vision Computing New Zealand,* Wellington, 2009, pp. 153-158.